\documentclass[letterpaper, 9 pt, conference]{ieeeconf}  

\IEEEoverridecommandlockouts                              

\usepackage{graphics} 
\usepackage{epsfig} 
\usepackage{mathptmx} 
\usepackage{times} 
\usepackage{amsmath} 
\usepackage{amssymb} 

\usepackage{booktabs}
\usepackage{multirow}
\usepackage[table,xcdraw]{xcolor}
\usepackage{calrsfs}
\DeclareMathAlphabet{\pazocal}{OMS}{zplm}{m}{n}

\makeatletter
\def\myitemize{%
    \ifnum\@itemdepth>3 \@toodeep\else
      \advance\@itemdepth\@ne
      \edef\@itemitem{labelitem\romannumeral\the\@itemdepth}%
      \list{\csname\@itemitem\endcsname}{\leftmargin=1.5em \labelwidth=1em \labelsep=0.5em \itemindent=0pt \parsep=0pt}%
    \fi}

\makeatother

\title{\LARGE \bf
BadHMP: Backdoor Attack Against Human Motion Prediction
}

\author{
        Chaohui Xu, Si Wang and Chip-Hong Chang\\
        School of Electrical and Electronic Engineering, Nanyang Technological University, Singapore
}

\begin{document}

\maketitle
\thispagestyle{empty}
\pagestyle{empty}

\begin{abstract}

Precise future human motion prediction over sub-second horizons from past observations is crucial for various safety-critical applications. To date, only a few studies have examined the vulnerability of skeleton-based neural networks to evasion and backdoor attacks. In this paper, we propose BadHMP, a novel backdoor attack that targets specifically human motion prediction tasks. Our approach involves generating poisoned training samples by embedding a localized backdoor trigger in one limb of the skeleton, causing selected joints to follow predefined motion in historical time steps. Subsequently, the future sequences are globally modified that all the joints move following the target trajectories. Our carefully designed backdoor triggers and targets guarantee the smoothness and naturalness of the poisoned samples, making them stealthy enough to evade detection by the model trainer while keeping the poisoned model unobtrusive in terms of prediction fidelity to untainted sequences. The target sequences can be successfully activated by the designed input sequences even with a low poisoned sample injection ratio. Experimental results on two datasets (Human3.6M and CMU-Mocap) and two network architectures (LTD and HRI) demonstrate the high-fidelity, effectiveness, and stealthiness of BadHMP. Robustness of our attack against fine-tuning defense is also verified.

\end{abstract}

\section{INTRODUCTION}

Human motion prediction is a sequence-to-sequence task where future motion sequences are predicted based on observed historical motion sequences. Accurate forecasting of future human poses is crucial for the success of various applications, such as human-robot interaction and collaboration (HRI/C)~\cite{moon2021fast,koppula2013anticipating}, human tracking~\cite{gong2011multi}, autonomous driving~\cite{paden2016survey}, and particularly in healthcare and biomedical fields, such as seamless interactions with exoskeletons and prosthetic devices that enable more effective rehabilitation~\cite{zhang2015real}.

Various advanced neural network architectures have been explored for this task, including recurrent neural networks (RNNs)~\cite{fragkiadaki2015recurrent,jain2016structural,martinez2017human}, graph convolutional networks (GCNs)~\cite{mao2019learning,mao2020history,zhong2022spatio}, generative models~\cite{komura2017recurrent,barsoum2018hp,yuan2020dlow,barquero2023belfusion}, and Transformers~\cite{aksan2021spatio,cai2020learning}. Despite the extensive research into deep learning based human motion prediction, the vulnerability of these models to potential attacks has not been sufficiently explored. To date, only a few evasion attacks~\cite{duan2024physics, medina2024fooling} have been investigated on human motion prediction. Hence, there exists a significant gap in understanding the robustness of human motion prediction models against other forms of malicious attacks.

Backdoor attack targeting deep neural networks (DNNs) is a form of data-poisoning attack, where the adversary subtly alters a small subset of training samples by embedding a trigger into the input data and substituting the corresponding outputs with predefined targets. During training, the victim model inadvertently learns both the intended tasks and a strong association between the trigger and the target output. At the inference stage, the model behaves as a benign model under normal conditions but consistently produces the predefined target output when the trigger is present in the input.

Most existing backdoor attacks focus on image classification tasks~\cite{gu2019badnets,nguyen2021wanet,li2019invisible,liu2020reflection,xu2023imperceptible}, while some studies have been extended to other tasks~\cite{cai2024towards,ye2022drinet,zhang2021backdoor,xi2021graph,sun2022backdoor}. While backdoor attacks have been studied in other skeleton-data-based machine leaning tasks~\cite{zhang2024invisibility, zheng2024towards}, they have not been explored in the context of human motion prediction models. A successful backdoor attack in this domain poses safety hazards that may lead to grave consequences. For instance, in the scenario of HRI/C, a robot equipped with a poisoned model may inaccurately predict human motions, leading to erroneous decisions and potentially hazardous outcomes in subsequent time steps. The main challenges of launching a backdoor attack on human motion prediction are as follows: 1) Due to the unique data format of human motion samples (spatial and temporal 3D joint positions), existing data-poisoning techniques are not directly applicable for generating the poisoned samples for such task; 2) To avoid detection by the model trainer, the poisoned training samples of human motion sequences must remain smooth and natural. This means that the clean samples need to be subtly manipulated to ensure that the fundamental physics principles of human body dynamics are not violated.

In this paper, we propose a novel backdoor attack to human motion prediction task dubbed $\textbf{BadHMP}$. The main contributions of our work are summarized as follows:

\begin{myitemize}
\item We propose a novel poisoning strategy that generates smooth and natural adversarial human motion samples. Specifically, we extract the motion of a selected limb from a source sample, and graft this predefined motion onto clean samples to seamlessly embed the backdoor trigger into input sequences. For output sequences, we globally extract and transfer the trajectories of all joints from the source sample to clean samples as the target motion patterns.
\item We design two novel evaluation metrics, Clean Data Error (CDE) and Backdoor Data Error (BDE), to assess the attack performance on human motion predictors.
\item Extensive experiments are conducted on two popular benchmark datasets (Human3.6M and CMU-Mocap) and two widely used model architectures (LTD and HRI) to attest the performance of our proposed attack.
\end{myitemize}

\section{Related Works}

\subsection{Human Motion Prediction}

Due to their good performance in sequence-to-sequence prediction tasks, RNNs have been extensively studied for human motion prediction. In the first RNN-based approach~\cite{fragkiadaki2015recurrent}, an Encoder-Recurrent-Decoder model was used for motion prediction. Subsequently, a Structural-RNN~\cite{jain2016structural} was developed to manually encode the spatial and temporal structures. To achieve multi-action predictions using a single model, a residual architecture for velocity prediction was proposed in~\cite{martinez2017human}. Numerous RNN-based methods~\cite{chiu2019action,corona2020context,wolter2018complex,gui2018adversarial} have since emerged, aiming to further enhance the prediction performance.

The GCN-based motion prediction method was first introduced in~\cite{mao2019learning} by employing the Discrete Cosine Transform (DCT) to encode the spatial dependency and temporal information of human poses. This approach was further refined by capturing similarities between current and historical motion contexts~\cite{mao2020history}. Inspired by the success of GCN in modeling dynamic relations among pose joints~\cite{mao2019learning}, various GCN-based prediction methods~\cite{zhong2022spatio,dang2021msr} have been developed for more complex spatio-temporal dependencies over diverse action types. Recently, the attention mechanism~\cite{aksan2021spatio,cai2020learning} of Transformers~\cite{vaswani2017attention} has been leveraged to capture the spatial, temporal, and pairwise joint relationships within motion sequences. Without resorting to complex deep learning architectures, a lightweight multi-layer perception combined with DCT and standard optimization techniques can achieve excellent performance with fewer parameters~\cite{guo2023back}. Sampling from deep generative models~\cite{komura2017recurrent,barsoum2018hp,yuan2020dlow,barquero2023belfusion} trained over large motion-capture dataset has also been devised for more realistic and coherent stochastic human motion prediction.

\subsection{Backdoor Attacks and Defenses}

Backdoor attacks train the victim model with poisoned training data to embed a malicious backdoor that can be activated at test time to cause the victim model to misbehave. The most popular attack dates back to BadNets~\cite{gu2019badnets}, where a small number of training samples are stamped with a tiny fixed binary pattern (a.k.a trigger) at the right bottom corner and relabeled to a target class. The backdoor feature is learnt by the DNN classifier during the training process. Since then, a series of improved backdoor attacks have been developed, employing techniques such as image blending transformations~\cite{chen2017targeted}, steganography~\cite{li2019invisible}, warping transformations~\cite{nguyen2021wanet}, and adaptive optimizations~\cite{nguyen2020input}, to enhance the stealthiness of the backdoor trigger, thereby evading detection by the model trainer. An even more inconspicuous branch of clean-label backdoor attacks~\cite{xu2023imperceptible,liu2020reflection,turner2018clean} can achieve target misclassification by hiding the backdoor triggers into the training images without altering their labels. While most existing backdoor attacks have been developed for image classification tasks, some successful attacks have also been reported in other task domains, such as speech recognition~\cite{cai2024towards,ye2022drinet}, graph classification~\cite{zhang2021backdoor,xi2021graph}, skeleton action recognition~\cite{zheng2024towards}, and human pose estimation~\cite{zhang2024invisibility} .

Defenses against backdoor attacks on computer vision and natural language processing models can be broadly categorized into two groups: detection and mitigation. Detection methods~\cite{wang2019neural,chou2020sentinet,chen2019deepinspect} focus on identifying whether the training dataset or the trained model has been embedded with a backdoor, while mitigation techniques~\cite{wu2021adversarial,li2021anti,doan2020februus,liu2018fine} aim to purify the poisoned training dataset or sanitize the victim model to reduce the success rate of backdoor activation by the triggered samples without compromising the prediction accuracy of benign samples.

\section{Threat Model}

Given a history motion sequence $X_{1:N} = \left[X_1, X_2, \cdots, X_N\right]$ that is composed of $N$ consecutive frames of human poses, the human motion prediction model $f_{\theta}$ parameterized by $\theta$ aims to forecast the future $T$ frames of poses as $X_{N+1:N+T} = \left[X_{N+1}, X_{N+2}, \cdots, X_{N+T}\right]$, where each pose $X_i \in \mathbb{R}^{K \times 3}$ consists of 3D coordinates of $K$ joints. Let $D_{\text{tr}}$ and ${D}_{\text{ts}}$ denote the training and test datasets, respectively. The training process aims to solve the following optimization problem:
\begingroup
\small
\begin{equation}
\begin{aligned}
        \theta^{\ast} & = \mathop{\arg\min} \mathbb{E}_{X \sim D_{\text{tr}}} \left[\pazocal{L}_{m}(\hat{X}_{N+1:N+T}, X_{N+1:N+T})\right], \\
                      & = \mathop{\arg\min} \mathbb{E}_{X \sim D_{\text{tr}}} \left[\frac{1}{K \times T} \sum_{n=1}^{T} \sum_{j=1}^K \left\|\hat{X}_{(j, n)}-X_{(j, n)}\right\|^2\right],
\end{aligned}
\end{equation}
\endgroup
where $\hat{X}_{N+1:N+T} = f_\theta(X_{1:N})$ and $X_{N+1:N+T}$ denote the predicted and ground-truth poses of future $T$ time steps, respectively. $\hat{X}_{(j, n)}$ represents the predicted $j$-th joint position at frame $n$, and $X_{(j, n)}$ is its corresponding ground truth. $\pazocal{L}_{m}$ denotes the Mean Per Joint Position Error (MPJPE)~\cite{ionescu2013human3} in millimeter, which is the most widely used metric for 3D pose error evaluation.

\subsection{Attack Scenario}

Following the threat model of backdoor attacks on image classification models~\cite{gu2019badnets,nguyen2021wanet,xu2023imperceptible}, we assume that the attacker is a malicious third party who provides the training set to the model trainer. In this scenario, the attacker is allowed to poison $\rho$\% of samples of the clean training set $D_\text{tr}$ before the training stage, where $\rho = (N_\text{poison}/N_\text{train})\times100\%$ is commonly referred to as the injection ratio. However, the attacker has no knowledge of or access to the training pipeline, including the model architecture, optimization algorithm, training loss, etc. Additionally, manipulating the well-trained victim model is also not permitted. The backdoor sample generation process can be expressed as $\tilde{X}=G(X)$, where $G(\cdot)$ denotes the poisoning function that will be elaborated in Sec.~\ref{sec:the_proposed_attack}, and $\tilde{X}$ is the poisoned sample.

\subsection{Attacker's Goals}

The attacker aims to launch a high-fidelity, effective and stealthy backdoor attack on the victim model $f_{\theta^{\prime}}$, with $\theta^{\prime}$ being the poisoned parameters.

\textbf{Fidelity.} The victim model $f_{\theta^{\prime}}$ is expected to perform normally as a benign model $f_\theta$ when fed with clean test samples to prevent the backdoor attack from being noticed by the model trainer. To assess this, we define a Clean Data Error (CDE) metric as follows:
\begingroup
\begin{equation}
        \text{CDE}(f) = \mathbb{E}_{X \sim D_{\text{ts}}} \left[\pazocal{L}_{m}(f(X_{1:N}), X_{N+1:N+T})\right].
\end{equation}
\endgroup

The CDE of the victim model should be comparable to that of the benign model, i.e., $\lvert\text{CDE}(f_{\theta^{\prime}}) - \text{CDE}(f_\theta)\rvert \leq \epsilon$, with $\epsilon$ being a small positive threshold.

\textbf{Effectiveness.} The victim model should produce incorrect sequences dictated by the attacker on triggered inputs at test time. We define the Backdoor Data Error (BDE) metric to evaluate the effectiveness of backdoor activation as:
\begingroup
\begin{equation}
        \text{BDE}(f) = \mathbb{E}_{\tilde{X} \sim \tilde{D}_{\text{ts}}} \left[\pazocal{L}_{m}(f(\tilde{X}_{1:N}), \tilde{X}_{N+1:N+T})\right],
\end{equation}
\endgroup
where $\tilde{D}_{\text{ts}}$ represents the poisoned test dataset in which all the samples are generated by $G(\cdot)$. A low value of $\text{BDE}(f_{\theta^{\prime}})$ implies that the victim model exhibits the incorrect behaviors expected by the attacker with high probability, thereby achieving a high attack success rate.

\begin{figure*}[t]
        \centering
        \includegraphics[width=0.95\linewidth]{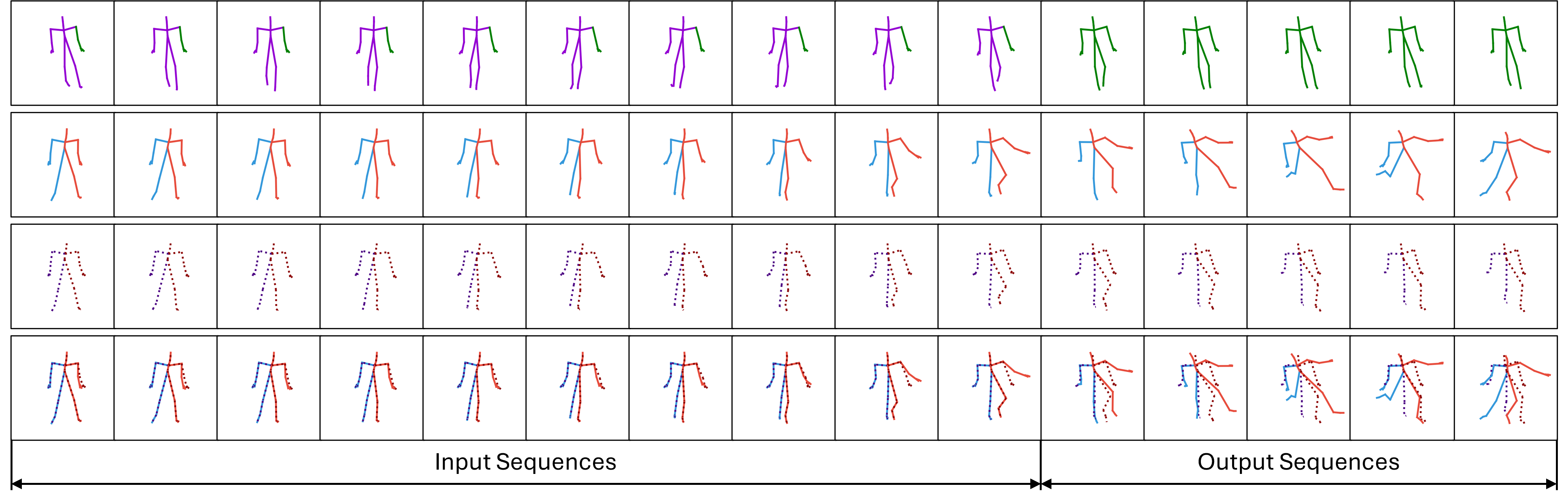}
        \caption{Row 1: the source sample with the semantic meaning of ``walking''. Only joints in green are leveraged to generate the trigger or target. Row 2: a clean sample with the semantic meaning of ``soccer''. Row 3: the poisoned version of the above clean sample. Row 4: Comparison of the paired clean (solid) and poisoned (dotted) samples. Due to the page limit, the complete 75-frame motion sample is down-sampled to 15 frames for display in this figure, with 10 frames for input and 5 frames for output.}
        \label{fig:visualization}
        \vspace{-4mm}
\end{figure*}

\textbf{Stealthiness.} Poisoned samples should look similar to its clean version to avoid being detected by human inspector or automatic checker. Specifically, the poisoned pose sequences should be \textbf{smooth} and \textbf{natural}. Mean per-joint acceleration (Acc) and jerk (second- and third-order derivatives of the joint positions, respectively) for human motion synthesis~\cite{yang2023qpgesture,rempe2021humor} are utilized to evaluate the smoothness of the poisoned samples. Moreover, following the physics-constrained attack~\cite{duan2024physics}, we compute the change of bone length to evaluate the naturalness. There exists a bone between a pair of connected joints (e.g., the humerus bone between shoulder and elbow), and the bone length change is small during the motion as human bones are not elastic. Hence, the bone length change (BLC) of poisoned samples should be kept as low as that of their clean versions. The above three metrics are formulated as follows:
\begingroup
\small
\begin{align}
        \text{Acc}(\textbf{X})  & = \frac{1}{K \times (N+T-2)} \sum_{n=1}^{N+T-2} \sum_{j=1}^K \left\|\ddot{\textbf{X}}\right\|^2_{(j,n)}, \\
        \text{Jerk}(\textbf{X}) & = \frac{1}{K \times (N+T-3)} \sum_{n=1}^{N+T-3} \sum_{j=1}^K \left\|\overset{\text{...}}{\textbf{X}}\right\|^2_{(j,n)}, \\
        \text{BLC}(\textbf{X})  & = \frac{1}{L_C \times (N+T-1)} \sum_{n=1}^{N+T-1} \sum_{l=1}^{L_C} \lvert S_{l, n+1} - S_{l, n} \rvert,
\end{align}
\endgroup
where $S_{l, n}$ denotes the length of the $l$-th bone at frame $n$, and $L_C$ is the total number of bones.

\section{The Proposed Attack} \label{sec:the_proposed_attack}

Our attack consists of three stages: (1) localized history sequences poisoning, (2) global future sequences poisoning, and (3) victim model poisoning.

\subsection{Localized History Sequences Poisoning}

A body pose is composed of five parts: torso, left arm, right arm, left leg, and right leg. On the $N$-frame input sequences, only several connected joints in a selected limb are manipulated for backdoor trigger embedding, while the remaining joints' positions are unchanged. This is to ensure that the semantic meaning of the input sequences is not damaged.

By default, we select the \textbf{left arm} for poisoning. Given a clean sample $X$ to be poisoned, and a source sample $X^{\text{src}}$ randomly selected from the clean test set. For the $N$ input frames, we graft the left arm pose of $X^{\text{src}}$ onto that of $X$. For instance, a ``$\textit{walking}$'' sample can be poisoned by controlling the $\textbf{left arm}$ to do a predefined motion grafted from the source sample like ``drawing a circle". However, since the clean and source samples may captured from actors of different physiques, direct grafting may causing disproportionate grafted left arm. Therefore, a 3D scaling transformation is first applied to $X^{\text{src}}$, producing $\overline{X}^\text{src} = \text{Scale}(X^\text{src}, X)$, where the bone lengths match those in $X$ but the directions remain the same as in $X^{\text{src}}$.

The left arm to be poisoned is formed by $m$ joints denoted as $J_p = \{j_1, j_2, \cdots, j_m\}$, and the joint of the left shoulder is $j_0$. The grafting operation is performed as:

\begingroup
\small
\begin{equation}
        \tilde{X}_{(j, n)}=
        \begin{cases}
                X_{(j_0, n)} + \left[ \overline{X}^\text{src}_{(j, n)} - \overline{X}^\text{src}_{(j_0, n)} \right], \qquad & \text{if} \  j \in J_p,\\
                X_{(j, n)},                                        & \text{otherwise},
        \end{cases}
        \label{eq:rigid_trigger}
\end{equation}
\endgroup
where $1 \leq n \leq N$. This trigger embedding process can be regarded as adopting a predefined left arm motion to the clean training input sequences without changing the bone lengths. Since the left arm motion is extracted from a real source sample, its smoothness and naturalness are guaranteed.

\subsection{Global Future Sequences Poisoning}

On the contrary, when poisoning the future sequences, we globally modify the 3D coordinates of all the $K$ joints by applying predefined trajectories. We first extract trajectories of all joints from the future sequences of the source sample, and subsequently add these trajectories to the last frame of the poisoned input sequences to generate the moving target as follows:
\begingroup
\small
\begin{equation}
        \tilde{X}_{(j, n)} = \tilde{X}_{(j, N)} + \left[ \overline{X}^{\text{src}}_{(j, n)} - \overline{X}^{\text{src}}_{(j, N)} \right], \  j \in \{1, 2, \cdots, K\},
        \label{eq:soft_target}
\end{equation}
\endgroup
where $N+1 \leq n \leq N+T$. Note that the source sample used for trajectories extraction can be either the same as or different from the one used for trigger embedding.

Fig.~\ref{fig:visualization} visualizes the source, clean, and poisoned examples to explain the whole poisoning process. The first row represents the source sample which is randomly selected from the test set of ``walking''. For the input sequences, only joints on the left arm (in green) are utilized to generated the trigger. For the output sequences, we compute the trajectories of all the $K$ joints as the target. The second and the third rows represent a clean sample of ``soccer'' and its poisoned version, respectively. To facilitate visual comparison, we jointly plot clean and poisoned samples in the last row. In the input time, the backdoor trigger is embedded only in the selected joints on the left arm, resulting in the overlapping of most parts of the clean and poisoned skeletons. Meanwhile, the semantic meaning of ``soccer'' is not damaged as we only slightly change the left arm motion of the poisoned sample. In the output time, the poisoned sequences exhibit large differences from the clean sequences because all the $K$ joints are applied with predefined trajectories extracted from the ``walking'' source sample.

\subsection{Victim Model Poisoning}

The attacker randomly selects a subset of clean samples from the training dataset and poison them. These poisoned samples can be either used to train a malicious model or released publicly as a contaminated dataset. During training, the victim model inadvertently learns a spurious correlation between the trigger (left arm motion) and the target trajectories. Consequently, when the trigger is detected during inference, the model fails to generate semantically correct predictions and instead outputting adversarial trajectories predefined by the attacker.

\begin{table*}[t]
        \caption{Action-wise prediction performance of the benign and victim LTD models on the H3.6M dataset.}
        \centering
        \renewcommand\arraystretch{0.6}

        \resizebox{0.90\linewidth}{!}{
                \begin{tabular}{@{\hspace{2mm}}c|l|cccc|cccc|cccc|cccc@{\hspace{2mm}}}
                        \toprule
                        Model                       & Time (ms)                      & 80          & 400          & 560          & 1000        & 80           & 400           & 560         & 1000       & 80           & 400           & 560           & 1000          & 80           & 400          & 560         & 1000        \\ \midrule
                                                    & \cellcolor[HTML]{C0C0C0}Action & \multicolumn{4}{c|}{\cellcolor[HTML]{C0C0C0}walking}    & \multicolumn{4}{c|}{\cellcolor[HTML]{C0C0C0}eating}     & \multicolumn{4}{c|}{\cellcolor[HTML]{C0C0C0}smoking}         & \multicolumn{4}{c}{\cellcolor[HTML]{C0C0C0}discussion}  \\ \cmidrule(l){2-18} 
                                                    & CDE                            & 11.5        & 41.6         & 46.7         & 51.1        & 8.1          & 37.2          & 49.0        & 71.3       & 8.1          & 38.6          & 49.6          & 71.6          & 12.7         & 67.8         & 86.4        & 121.7       \\
                                                    & BDE                            & 39.5        & 135.6        & 167.9        & 158.5       & 35.3         & 117.8         & 159.2       & 153.9      & 34.6         & 112.5         & 154.3         & 152.7         & 38.3         & 123.3        & 172.1       & 170.8       \\ \cmidrule(l){2-18} 
                                                    & \cellcolor[HTML]{C0C0C0}Action & \multicolumn{4}{c|}{\cellcolor[HTML]{C0C0C0}directions} & \multicolumn{4}{c|}{\cellcolor[HTML]{C0C0C0}greeting}   & \multicolumn{4}{c|}{\cellcolor[HTML]{C0C0C0}phoning}         & \multicolumn{4}{c}{\cellcolor[HTML]{C0C0C0}posing}      \\ \cmidrule(l){2-18} 
                                                    & CDE                            & 9.0         & 59.2         & 81.4         & 119.2       & 17.0         & 84.8          & 105.9       & 137.7      & 10.1         & 52.1          & 69.5          & 109.4         & 13.9         & 86.8         & 119.9       & 181.9       \\
                                                    & BDE                            & 36.5        & 118.4        & 170.6        & 170.8       & 38.8         & 123.1         & 169.4       & 166.0      & 36.1         & 113.6         & 154.4         & 151.1         & 38.8         & 128.5        & 180.0       & 188.6       \\ \cmidrule(l){2-18} 
                                                    & \cellcolor[HTML]{C0C0C0}Action & \multicolumn{4}{c|}{\cellcolor[HTML]{C0C0C0}purchases}  & \multicolumn{4}{c|}{\cellcolor[HTML]{C0C0C0}sitting}    & \multicolumn{4}{c|}{\cellcolor[HTML]{C0C0C0}sittingdown}     & \multicolumn{4}{c}{\cellcolor[HTML]{C0C0C0}takingphoto} \\ \cmidrule(l){2-18} 
                                                    & CDE                            & 14.3        & 73.1         & 97.1         & 132.5       & 10.1         & 57.0          & 79.2        & 132.0      & 16.7         & 77.8          & 105.6         & 163.3         & 9.8          & 60.0         & 86.0        & 146.9       \\
                                                    & BDE                            & 38.0        & 127.0        & 180.6        & 189.1       & 35.1         & 113.8         & 160.1       & 164.1      & 36.9         & 115.6         & 162.0         & 170.8         & 34.8         & 113.8        & 167.3       & 180.0       \\ \cmidrule(l){2-18} 
                                                    & \cellcolor[HTML]{C0C0C0}Action & \multicolumn{4}{c|}{\cellcolor[HTML]{C0C0C0}waiting}    & \multicolumn{4}{c|}{\cellcolor[HTML]{C0C0C0}walkingdog} & \multicolumn{4}{c|}{\cellcolor[HTML]{C0C0C0}walkingtogether} & \multicolumn{4}{c}{\cellcolor[HTML]{FFCCC9}average}     \\ \cmidrule(l){2-18} 
                                                    & CDE                            & 10.7        & 61.5         & 82.9         & 112.9       & 22.8         & 94.9          & 116.6       & 160.3      & 10.6         & 43.8          & 52.6          & 63.0          & 12.4         & 62.4         & 81.9        & 118.3       \\
                        \multirow{-12}{*}{benign}   & BDE                            & 36.1        & 116.2        & 161.0        & 161.4       & 39.4         & 130.3         & 178.7       & 188.6      & 38.0         & 126.1         & 163.3         & 160.8         & 37.1         & 121.1        & 166.7       & 168.5       \\ \midrule
                                                    & \cellcolor[HTML]{C0C0C0}Action & \multicolumn{4}{c|}{\cellcolor[HTML]{C0C0C0}walking}    & \multicolumn{4}{c|}{\cellcolor[HTML]{C0C0C0}eating}     & \multicolumn{4}{c|}{\cellcolor[HTML]{C0C0C0}smoking}         & \multicolumn{4}{c}{\cellcolor[HTML]{C0C0C0}discussion}  \\ \cmidrule(l){2-18} 
                                                    & CDE                            & 11.2        & 40.4         & 46.5         & 51.5        & 8.0          & 37.9          & 50.9        & 73.6       & 8.2          & 38.9          & 51.2          & 76.2          & 12.6         & 70.1         & 91.5        & 129.7       \\
                                                    & BDE                            & 3.1         & 4.3          & 8.3          & 6.3         & 2.9          & 4.3           & 8.9         & 6.1        & 2.9          & 4.2           & 8.6           & 6.7           & 3.0          & 4.4          & 8.9         & 6.6         \\ \cmidrule(l){2-18} 
                                                    & \cellcolor[HTML]{C0C0C0}Action & \multicolumn{4}{c|}{\cellcolor[HTML]{C0C0C0}directions} & \multicolumn{4}{c|}{\cellcolor[HTML]{C0C0C0}greeting}   & \multicolumn{4}{c|}{\cellcolor[HTML]{C0C0C0}phoning}         & \multicolumn{4}{c}{\cellcolor[HTML]{C0C0C0}posing}      \\ \cmidrule(l){2-18} 
                                                    & CDE                            & 8.6         & 57.5         & 80.3         & 115.4       & 16.6         & 84.3          & 106.4       & 142.4      & 9.9          & 51.9          & 70.3          & 111.2         & 13.5         & 85.9         & 120.8       & 186.4       \\
                                                    & BDE                            & 2.9         & 4.0          & 8.3          & 5.9         & 3.1          & 4.3           & 8.8         & 6.3        & 3.0          & 4.6           & 9.1           & 7.2           & 3.1          & 5.0          & 9.4         & 7.4         \\ \cmidrule(l){2-18} 
                                                    & \cellcolor[HTML]{C0C0C0}Action & \multicolumn{4}{c|}{\cellcolor[HTML]{C0C0C0}purchases}  & \multicolumn{4}{c|}{\cellcolor[HTML]{C0C0C0}sitting}    & \multicolumn{4}{c|}{\cellcolor[HTML]{C0C0C0}sittingdown}     & \multicolumn{4}{c}{\cellcolor[HTML]{C0C0C0}takingphoto} \\ \cmidrule(l){2-18} 
                                                    & CDE                            & 14.1        & 73.8         & 100.0        & 138.6       & 10.1         & 56.4          & 78.5        & 131.1      & 16.5         & 76.1          & 103.6         & 160.5         & 9.5          & 58.9         & 87.8        & 151.0       \\
                                                    & BDE                            & 3.1         & 5.1          & 9.4          & 7.6         & 3.1          & 5.7           & 10.1        & 9.1        & 3.2          & 6.5           & 10.8          & 9.5           & 3.0          & 5.3          & 9.6         & 7.6         \\ \cmidrule(l){2-18} 
                                                    & \cellcolor[HTML]{C0C0C0}Action & \multicolumn{4}{c|}{\cellcolor[HTML]{C0C0C0}waiting}    & \multicolumn{4}{c|}{\cellcolor[HTML]{C0C0C0}walkingdog} & \multicolumn{4}{c|}{\cellcolor[HTML]{C0C0C0}walkingtogether} & \multicolumn{4}{c}{\cellcolor[HTML]{FFCCC9}average}     \\ \cmidrule(l){2-18} 
                                                    & CDE                            & 10.5        & 61.6         & 83.8         & 114.8       & 21.9         & 95.0          & 119.2       & 171.6      & 10.6         & 43.8          & 53.6          & 62.7          & 12.1         & 62.2         & 83.0        & 121.1       \\
                        \multirow{-12}{*}{victim}   & BDE                            & 2.9         & 4.4          & 8.8          & 6.5         & 3.3          & 5.5           & 9.6         & 8.2        & 3.2          & 4.5           & 9.3           & 6.1           & 3.1          & 4.8          & 9.2         & 7.1         \\ \bottomrule
                \end{tabular}
        }
        \label{tab:action_wise}
        \vspace{-4mm}
\end{table*}

\section{Experiments and Results}

\subsection{Experimental Settings}

\subsubsection{Datasets}

The proposed backdoor attack is evaluated on two benchmark motion capture (mocap) datasets, including Human3.6M (H3.6M)~\cite{ionescu2013human3} and CMU-Mocap (CMU) \textsuperscript{\ref{note1}}.

\footnotetext[1]{\label{note1} http://mocap.cs.cmu.edu}

\textbf{H3.6M} is the most widely used large-scale dataset for human motion prediction, comprising 3.6 million 3D human poses. It includes motion sequences of 7 actors performing 15 distinct actions. The human skeleton is composed of 32 joints expressed by exponential maps. We convert these representations to 3D coordinates and use the remaining 22 joints after removing 10 redundant joints.

\textbf{CMU} contains 8 categories of actions where the 38-joint skeletons are also originally presented by exponential maps. Like H3.6M, these presentations are converted to 3D coordinates, and this dataset is evaluated on 25 joints.

For both two datasets, the samples are divided to training and test sets following the configuration of~\cite{mao2019learning}. To balance different actions with different sequence lengths and avoid high variance, we take 256 random samples per action for testing as in~\cite{mao2020history,guo2023back}.

\subsubsection{Model Architectures}

Our attack is evaluated on two model architectures: LearningTrajectoryDependency (LTD)~\cite{mao2019learning} and HistoryRepeatItself (HRI)~\cite{mao2020history}.

\subsubsection{Implementation Details}

The model is trained to predict both short-term (0 to 500 ms) and long-term (500 to 1000 ms) future human motions. The input length $N$ and the output length $T$ are set to 50 and 25, respectively. The default injection ratio $\rho$ is 10\%. We use the Adam optimizer and a batch size of 256 for training. The network is trained for 50 epochs, with the learning rate initially set to 0.01 and decayed by a factor of 0.96 every two epochs. For evaluation, we measure the CDE and BDE of the model at 80 and 400 ms for short-term prediction, and 560 and 1000 ms for long-term prediction.

\begin{table}[t]
        \caption{Averaged CDE and BDE measured on the H3.6M dataset.}
        \centering
        \renewcommand\arraystretch{0.6}

        \resizebox{0.98\linewidth}{!}{
                \begin{tabular}{@{\hspace{2mm}}c|lcccc|ccccc@{\hspace{2mm}}}
                \toprule
                                        & \multicolumn{5}{c|}{LTD}                                      & \multicolumn{5}{c}{HRI}                                       \\ \midrule
                Model                   & \multicolumn{1}{l|}{Time (ms)} & 80   & 400   & 560   & 1000  & \multicolumn{1}{c|}{Time (ms)} & 80   & 400   & 560   & 1000  \\ \midrule
                \multirow{2}{*}{benign} & \multicolumn{1}{l|}{CDE}       & 12.4 & 62.4  & 81.9  & 118.3 & \multicolumn{1}{c|}{CDE}       & 11.9 & 62.9  & 84.5  & 123.9 \\
                                        & \multicolumn{1}{l|}{BDE}       & 37.1 & 121.0 & 166.7 & 168.5 & \multicolumn{1}{c|}{BDE}       & 36.7 & 119.9 & 168.2 & 170.3 \\ \midrule
                \multirow{2}{*}{victim} & \multicolumn{1}{l|}{CDE}       & 12.1 & 62.2  & 83.0  & 121.1 & \multicolumn{1}{c|}{CDE}       & 12.2 & 63.0  & 83.7  & 121.1 \\
                                        & \multicolumn{1}{l|}{BDE}       & 3.1  & 4.8   & 9.2   & 7.1   & \multicolumn{1}{c|}{BDE}       & 2.7  & 5.3   & 6.7   & 7.4   \\ \bottomrule
                \end{tabular}
        }
        \label{tab:average_h36m}
        \vspace{-3mm}
\end{table}

\begin{table}[t]
        \caption{Averaged CDE and BDE measured on the CMU dataset.}
        \centering
        \renewcommand\arraystretch{0.6}

        \resizebox{0.98\linewidth}{!}{
                \begin{tabular}{@{\hspace{2mm}}c|lcccc|ccccc@{\hspace{2mm}}}
                \toprule
                                        & \multicolumn{5}{c|}{LTD}                                      & \multicolumn{5}{c}{HRI}                                       \\ \midrule
                Model                   & \multicolumn{1}{l|}{Time (ms)} & 80   & 400   & 560   & 1000  & \multicolumn{1}{c|}{Time (ms)} & 80   & 400   & 560   & 1000  \\ \midrule
                \multirow{2}{*}{benign} & \multicolumn{1}{l|}{CDE}       & 10.8 & 44.4  & 59.6  & 89.8  & \multicolumn{1}{l|}{CDE}       & 10.6 & 43.0  & 58.0  & 89.1  \\
                                        & \multicolumn{1}{l|}{BDE}       & 25.4 & 120.2 & 166.1 & 231.5 & \multicolumn{1}{l|}{BDE}       & 25.6 & 122.3 & 170.3 & 239.5 \\ \midrule
                \multirow{2}{*}{victim} & \multicolumn{1}{l|}{CDE}       & 10.9 & 44.1  & 58.3  & 88.9  & \multicolumn{1}{l|}{CDE}       & 10.7 & 43.4  & 57.6  & 87.3  \\
                                        & \multicolumn{1}{l|}{BDE}       & 6.4  & 5.0   & 5.9   & 8.9   & \multicolumn{1}{l|}{BDE}       & 5.4  & 4.1   & 5.8   & 7.4   \\ \bottomrule
                \end{tabular}
        }
        \label{tab:average_cmu}
        \vspace{-3mm}
\end{table}

\subsection{Evaluation}

\subsubsection{Fidelity and Effectiveness} To evaluate the fidelity of the poisoned model and the effectiveness of backdoor activation, we train a benign model and a victim model on the clean training dataset $D_\text{tr}$ and the poisoned training dataset $\tilde{D}_\text{tr}$, respectively, and measure their prediction performance on both clean and poisoned test sets. Table~\ref{tab:action_wise} reports the action-wise CDE and BDE of both benign and victim models, with average values in the red cells. The dataset used for the evaluation is H3.6M, and the model architecture is LTD. All poisoned training and test samples are generated by the same source sample.

The benign model demonstrates excellent prediction performance on clean test samples, with an average CDE of approximately 12 at 80 ms. As prediction time increases, the CDE gradually rises, which is expected because the prediction errors accumulate over time. In contrast, the benign model's BDE is significantly higher than the CDE because it does not learn the backdoor features during training. As a result, it fails to produce the attacker expected predictions when trigger-embedded input sequences are encountered.

The CDE of the victim model remains very close to that of the benign model across all actions and evaluation time steps. This indicates that the ``fidelity'' criterion is met, as the victim model behaves like a normally trained model when processing clean test samples. However, for test samples generated by the specific poisoning strategy, the BDE of the victim model is significantly lower compared to the CDE. Additionally, the BDE accumulates much more slowly than the CDE over time. Specifically, the victim model's average CDE and BDE are 12.1 and 3.1 at 80 ms, and 121.1 and 7.1 at 1000 ms, respectively. These results show that the embedded backdoor can be successfully activated at test time, causing the victim model to accurately produce the target sequences as intended by the attacker, even for long-term predictions. Thus, our attack also fulfills the effectiveness criterion.

The attack performances in other cases across various datasets and model architectures are also evaluated. Due to the page limit, we only report the average CDE and BDE of these experiments in Table~\ref{tab:average_h36m} and Table~\ref{tab:average_cmu}. For all the evaluated cases, the victim model achieves low CDE comparable to that of the benign model, indicating that its prediction performance on clean test samples are not weakened. Meanwhile, the BDE measured on victim models is consistently and significantly lower than the corresponding CDE, indicating that the attack is satisfactorily effective across all cases. In summary, all victim models trained on the poisoned dataset have high fidelity and the target sequences can be effectively activated, regardless of the datasets and model architectures. 

\begin{figure}[t]
        \centering
        \includegraphics[width=0.9\linewidth]{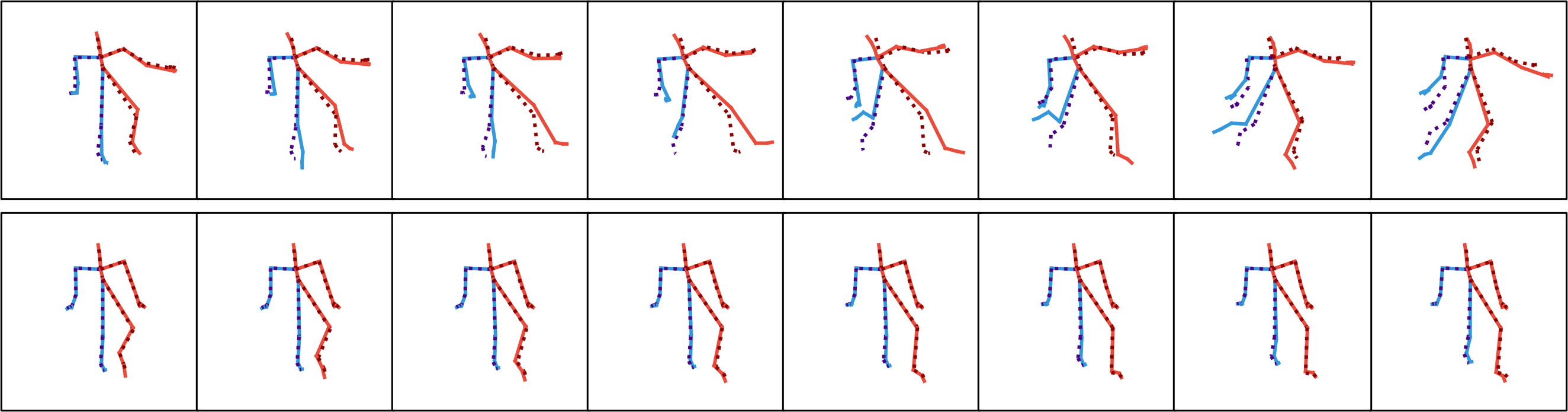}
        \caption{Visualization of a victim LTD model's predictions on clean and poisoned input sequences. Row 1: clean output sequences (solid) and the victim model's prediction on clean input sequences (dotted). Row 2: poisoned output sequences (solid) and the victim model's prediction on poisoned input sequences (dotted).}
        \label{fig:prediciton}
        \vspace{-4mm}
\end{figure}

Fig.~\ref{fig:prediciton} visualizes the behavior of a victim LTD model when presented with clean and poisoned input sequences. The original test sample corresponds to the action ``soccer''. When it is fed with clean test sequences, the victim model behaves as a benign model, with its output closely matching the ground-truth future sequences. However, when the input sequences are embedded with a backdoor trigger, the victim model produces incorrect predictions as intended by the attacker. The semantic meaning of the predicted future sequences is altered and the motion is no longer identified as ``soccer''. Notably, the prediction error accumulates more rapidly over time with clean input sequences. This observation aligns with the findings from the quantitative results provided in the tables. They demonstrate that the victim model exhibits lower BDE than CDE, and the embedded backdoor can be easily and effectively activated during inference.

\begin{table}[h]
        \caption{Maximum Acc, maximum jerk, and averaged BLC measured on paired clean and poisoned training samples.}
        \centering
        \renewcommand\arraystretch{0.6}

        \resizebox{0.85\linewidth}{!}{
                \begin{tabular}{@{\hspace{2mm}}l|cc|cc@{\hspace{2mm}}}
                \toprule
                Dataset   & \multicolumn{2}{c|}{H3.6M}            & \multicolumn{2}{c}{CMU}               \\ \midrule
                Metric    & \multicolumn{1}{c|}{clean} & poisoned & \multicolumn{1}{c|}{clean} & poisoned \\ \midrule
                max. Acc  & 44.14                      & 43.65    & 13.79                      & 13.61    \\
                max. Jerk & 72.66                      & 75.72    & 23.14                      & 24.58    \\
                avg. BLC  & 238.87                     & 239.08   & 337.19                     & 337.35   \\ \bottomrule
                \end{tabular}
        }
        \label{tab:stealthy_eval}
\end{table}

\subsubsection{Stealthiness} Table~\ref{tab:stealthy_eval} reports the maximum acceleration, maximum jerk, and averaged BLC measured on paired clean and poisoned samples.

It shows that the poisoned samples exhibit kinematically plausible motion patterns, with maximum acceleration and jerk values closely matching those of clean samples. This is a direct result of our carefully designed trigger and target, which enforce smooth spatiotemporal transitions. Meanwhile, by applying a bone-length-aware scaling transformation to the source sample before trigger and target extraction, we ensure the BLC of poisoned and clean sequences remains statistically indistinguishable. These results collectively validate the stealthiness of our attack under kinematic metrics.

\subsubsection{Effect of the Injection Ratio $\rho$} The default injection ratio is set to 10\% for all the experiments presented earlier. To investigate the effect of injection ratio on attack performance, we trained multiple victim models on datasets poisoned with different ratios, specifically $\rho \in \{2\%, 5\%, 8\%, 10\%, 15\%\}$. The results are presented in Table~\ref{tab:injection_ratio}, where $\rho = 0\%$ corresponds to the benign model. The results show that the CDE of poisoned models is insensitive to changes in $\rho$. Even at an injection ratio of 15\%, the CDE of the victim model remains very close to that of the benign model, thus further confirming that the victim model maintains a high fidelity even if it is heavily poisoned.

Moreover, the BDE gradually decreases as $\rho$ increases, which is expected because a higher proportion of poisoned training samples allows the victim model to better learn the association between the trigger and the target. The 10\% default injection ratio provides a good balance of effectiveness (low BDE) and stealthiness (low injection ratio) on both datasets.

\begin{table}[t]
        \caption{Attack performance under various injection ratios.}
        \centering
        \renewcommand\arraystretch{0.6}

        \resizebox{1.0\linewidth}{!}{
                \begin{tabular}{@{\hspace{2mm}}clcccc|clcccc@{\hspace{2mm}}}
                        \toprule
                        \multicolumn{6}{c|}{H3.6M dataset, LTD model}                                              & \multicolumn{6}{c}{CMU dataset, HRI model}                                                 \\ \midrule
                        \multicolumn{1}{c|}{$\rho$ (\%)}         & \multicolumn{1}{l|}{Time (ms)} & 80   & 400   & 560   & 1000  & \multicolumn{1}{c|}{$\rho$ (\%)}         & \multicolumn{1}{l|}{Time (ms)} & 80   & 400   & 560   & 1000  \\ \midrule
                        \multicolumn{1}{c|}{\multirow{2}{*}{0}}  & \multicolumn{1}{l|}{CDE}       & 12.4 & 62.4  & 81.9  & 118.3 & \multicolumn{1}{c|}{\multirow{2}{*}{0}}  & \multicolumn{1}{l|}{CDE}       & 10.6 & 43.0  & 58.0  & 89.1  \\
                        \multicolumn{1}{c|}{}                    & \multicolumn{1}{l|}{BDE}       & 37.1 & 121.0 & 166.7 & 168.5 & \multicolumn{1}{c|}{}                    & \multicolumn{1}{l|}{BDE}       & 25.6 & 122.3 & 170.3 & 239.5 \\ \midrule
                        \multicolumn{1}{c|}{\multirow{2}{*}{2}}  & \multicolumn{1}{l|}{CDE}       & 12.6 & 62.0  & 81.2  & 115.4 & \multicolumn{1}{c|}{\multirow{2}{*}{2}}  & \multicolumn{1}{l|}{CDE}       & 10.7 & 43.3  & 56.6  & 88.1  \\
                        \multicolumn{1}{c|}{}                    & \multicolumn{1}{l|}{BDE}       & 8.4  & 12.4  & 13.5  & 13.8  & \multicolumn{1}{c|}{}                    & \multicolumn{1}{l|}{BDE}       & 14.5 & 30.1  & 34.5  & 46.2  \\ \midrule
                        \multicolumn{1}{c|}{\multirow{2}{*}{5}}  & \multicolumn{1}{l|}{CDE}       & 12.2 & 62.6  & 83.3  & 120.5 & \multicolumn{1}{c|}{\multirow{2}{*}{5}}  & \multicolumn{1}{l|}{CDE}       & 10.5 & 42.6  & 56.1  & 88.6  \\
                        \multicolumn{1}{c|}{}                    & \multicolumn{1}{l|}{BDE}       & 5.5  & 5.9   & 7.2   & 9.2   & \multicolumn{1}{c|}{}                    & \multicolumn{1}{l|}{BDE}       & 8.0  & 8.9   & 11.0  & 16.7  \\ \midrule
                        \multicolumn{1}{c|}{\multirow{2}{*}{8}}  & \multicolumn{1}{l|}{CDE}       & 12.4 & 62.5  & 82.0  & 116.7 & \multicolumn{1}{c|}{\multirow{2}{*}{8}}  & \multicolumn{1}{l|}{CDE}       & 10.6 & 42.4  & 56.5  & 87.0  \\
                        \multicolumn{1}{c|}{}                    & \multicolumn{1}{l|}{BDE}       & 3.7  & 7.3   & 11.0  & 8.6   & \multicolumn{1}{c|}{}                    & \multicolumn{1}{l|}{BDE}       & 5.9  & 4.7   & 6.1   & 8.2   \\ \midrule
                        \multicolumn{1}{c|}{\multirow{2}{*}{10}} & \multicolumn{1}{l|}{CDE}       & 12.1 & 62.2  & 83.0  & 121.1 & \multicolumn{1}{c|}{\multirow{2}{*}{10}} & \multicolumn{1}{l|}{CDE}       & 10.7 & 43.4  & 57.6  & 87.3  \\
                        \multicolumn{1}{c|}{}                    & \multicolumn{1}{l|}{BDE}       & 3.1  & 4.8   & 9.2   & 7.1   & \multicolumn{1}{c|}{}                    & \multicolumn{1}{l|}{BDE}       & 5.4  & 4.1   & 5.8   & 7.4   \\ \midrule
                        \multicolumn{1}{c|}{\multirow{2}{*}{15}} & \multicolumn{1}{l|}{CDE}       & 12.4 & 62.6  & 82.5  & 118.3 & \multicolumn{1}{c|}{\multirow{2}{*}{15}} & \multicolumn{1}{l|}{CDE}       & 10.7 & 43.0  & 56.9  & 89.9  \\
                        \multicolumn{1}{c|}{}                    & \multicolumn{1}{l|}{BDE}       & 2.3  & 5.8   & 6.0   & 5.8   & \multicolumn{1}{c|}{}                    & \multicolumn{1}{l|}{BDE}       & 3.9  & 4.0   & 5.0   & 6.9   \\ \bottomrule
                \end{tabular}
        }
        \label{tab:injection_ratio}
        \vspace{-4mm}
\end{table}

\begin{table}[h]
        \caption{Robustness of BadHMP against fine-tuning defense.}
        \centering
        \renewcommand\arraystretch{0.6}

        \resizebox{1.0\linewidth}{!}{
                \begin{tabular}{@{\hspace{2mm}}clcccc|clcccc@{\hspace{2mm}}}
                        \toprule
                        \multicolumn{6}{c|}{H3.6M dataset, LTD model}                                              & \multicolumn{6}{c}{CMU dataset, HRI model}                                                \\ \midrule
                        \multicolumn{1}{c|}{}                        & \multicolumn{1}{l|}{Time (ms)} & 80   & 400  & 560  & 1000  & \multicolumn{1}{c|}{}                        & \multicolumn{1}{l|}{Time (ms)} & 80   & 400  & 560  & 1000 \\ \midrule
                        \multicolumn{1}{c|}{\multirow{2}{*}{Before}} & \multicolumn{1}{l|}{CDE}       & 12.1 & 62.2 & 83.0 & 121.1 & \multicolumn{1}{c|}{\multirow{2}{*}{Before}} & \multicolumn{1}{l|}{CDE}       & 10.7 & 43.4 & 57.6 & 87.3 \\
                        \multicolumn{1}{c|}{}                        & \multicolumn{1}{l|}{BDE}       & 3.1  & 4.8  & 9.2  & 7.1   & \multicolumn{1}{c|}{}                        & \multicolumn{1}{l|}{BDE}       & 5.4  & 4.1  & 5.8  & 7.4  \\ \midrule
                        \multicolumn{1}{c|}{\multirow{2}{*}{After}}  & \multicolumn{1}{l|}{CDE}       & 12.0 & 63.1 & 83.9 & 122.7 & \multicolumn{1}{c|}{\multirow{2}{*}{After}}  & \multicolumn{1}{l|}{CDE}       & 10.6 & 43.4 & 57.4 & 88.0 \\
                        \multicolumn{1}{c|}{}                        & \multicolumn{1}{l|}{BDE}       & 9.3  & 20.3 & 20.9 & 21.7  & \multicolumn{1}{c|}{}                        & \multicolumn{1}{l|}{BDE}       & 8.2  & 10.3 & 12.6 & 16.1 \\ \bottomrule
                \end{tabular}
        }
        \label{tab:ft_defense}
\end{table}

\subsubsection{Robustness} Most existing backdoor defenses~\cite{wang2019neural,chou2020sentinet,chen2019deepinspect,doan2020februus} designed for image classification tasks are not applicable to human motion prediction due to the differences in data format. To evaluate the robustness of our attack, we test its resilience against fine-tuning—a universal defense mechanism. The defender is assumed to retain 30\% of the original clean training samples and fine-tunes the victim model for 30 epochs. As shown in Table~\ref{tab:ft_defense}, although fine-tuning increases the BDE, the value remains significantly lower than the corresponding CDE. This demonstrates that the embedded backdoor retains its activation capability, confirming the robustness of our attack against fine-tuning defense.

\section{CONCLUSION}

This paper proposed $\textbf{BadHMP}$, a novel backdoor attack targeting human motion prediction tasks. Our key innovation lies in a two-stage poisoning strategy: 1) We extract the motion of a selected limb from a source sample and graft it onto historical input sequences of clean samples. 2) For future sequences, we globally transfer the joint trajectories from the source sample to clean samples. The poisoned samples exhibit provable stealthiness, as both trigger and target are derived from real-world data, inherently satisfying naturalness and accessibility criteria for human bodies. The prediction fidelity of the poisoned model to benign input sequences, the activation success rate of target sequences, and the smoothness and naturalness of the trigger sequences of BadHMP have been comprehensively evaluated by objective quantitative metrics on two datasets and two model architectures.

\bibliographystyle{IEEEtran}
\bibliography{ref}

\end{document}